\documentclass[runningheads]{llncs}
\usepackage{algorithm}
\usepackage{graphicx}
\usepackage{algorithmic}
\usepackage{amsmath,amssymb,amsfonts}
\usepackage{amsfonts}

\begin{document}
\title{Neural Architecture Search in Embedding Space}
%
%
\author{Chun Ting Liu}

%
\authorrunning{Chun Ting Liu }

%
\institute{Data Science Lab Team, Cathay Fincial Holdings, Taiwan \\
\email{Jimliu@cathayholdings.com.tw}}
%
\maketitle              

\begin{abstract}
The neural architecture search (NAS) algorithm with reinforcement learning can be a powerful and novel framework for the automatic discovering process of neural architectures. However, its application is restricted by noncontinuous and high-dimensional search spaces, which result in difficulty in optimization. To resolve these problems, we proposed NAS in embedding space (NASES), which is a novel framework. Unlike other NAS with reinforcement learning approaches that search over a discrete and high-dimensional architecture space, this approach enables reinforcement learning to search in an embedding space by using architecture encoders and decoders. The current experiment demonstrated that the performance of the final architecture network using the NASES procedure is comparable with that of other popular NAS approaches for the image classification task on CIFAR-10. The results of the experiment were efficient and indicated that NASES was highly efficient to discover final architecture only in $<$3.5 GPU hours. The beneficial-performance and effectiveness of NASES was impressive when the architecture-embedding searching and weight initialization were applied. 

\end{abstract}
\section{Introduction}
\noindent 
Deep neural networks have enabled advances in image recognition, sequential pattern recognition, recommendation systems, and various tasks in the past decades. However, selecting a suitable neural architecture is frequently arduous because of the classical and new neural architectures emerging daily. In general, manual design of network architectures according to the cases is achievable. However, hyperparameter tuning and architecture engineering through manual selection requires considerable time. Furthermore, manually designing a neural network architecture requires substantial experience in deep learning.

Given the aforementioned reasons, neural architecture search (NAS), an automated architecture engineering, has been successful in the past years. The NAS algorithm is divided into three dimensions, namely search space, search strategy, and performance estimation strategy \cite{Elsken2019}. Outstanding results have been achieved using NAS with the reinforcement learning search strategy \cite{zoph2017}. Here, a recurrent network was used to generate a string to form a child network. However, such a type of network exhibits two problems: noncontinuous and high-dimensional search space. The frequent large strings of action from the recurrent network and the discrete space result in difficulty in optimization. The critical contribution of this study is the improvement of the dimension and quality of the search space that could provide a more efficient framework to solve the two problems of searching architectures.

If a vector that can represent network architecture without discrete values is determined, then the noncontinuous aforementioned disadvantages can be addressed. We proposed the NAS in embedding space (NASES) method, which involves mapping origin architecture to architecture-embedding by using an architecture encoder. The advantage of embedding space includes the lower-dimensional and continues space, it considerably alleviates the difficulty in the optimization problem of the NAS procedure with reinforcement learning. To learn and search on the embedding space, we developed a mechanism to generate architecture encoder and decoder to promote origin architecture communication with the embedding space, and the autoencoder network was used in the mechanism \cite{hinton2006}. The architecture simulator simulates the origin architecture space, which assists the real architecture encoder learning. The decoder realizes the relationship between origin architecture and architecture-embedding, which maps the architecture-embedding to the origin architecture.

The NASES procedure was implemented in two stages. We obtained a pretraining architecture decoder and a pretraining architecture simulator in the first stage. In the second stage, we used the NASES procedure for image classification on CIFAR-10 by using weights initialization from the  pretrained network in the first stage. The results of the experiment were efficient and indicated that NASES was highly efficient in $<$ 3.5 GPU hours. Thus, the results were comparable with that of other popular NAS methods.

\section{Related Work}
\subsection{Reinforcement Learning with Action Embedding}
    Reinforcement learning is a general approach that can be applied broadly to various areas. However, the large and discrete action space causes problems in function approximation. The majority of studies have focused on two approaches, one approach factorizes the action space into binary subspaces \cite{pazis2011,dulac2012}. The other approach involves embedding discrete actions into a continuous action, determining optimal actions in the continuous space, and selecting the nearest discrete action to reduce the scaling of action sizes \cite{dulac2015,hasselt2009}.

\subsection{Search Strategy with Reinforcement Learning}
    Our search strategy is based on reinforcement learning. Zoph\cite{zoph2017} provided a novel NAS framework, which incorporated reinforcement learning and applied it to two agents of the child network and the controller of the recurrent network. The child network generated neural architecture that can be considered the action of the controller network. Unlike the use of policy gradient by Zoph\cite{zoph2017}, Will\cite{will1992}, and Baker\cite{baker2016} used q-learning to update the weight of the network.

\subsection{NAS with Continues Vector}
    Most NAS procedures use a discrete search space. Unlike other approaches, such as the learning over discrete and nondifferentiable search space, Liu\cite{liu2018} proposed an approach of differentiable architecture search (DARTS), which was based on the continuous relaxation of the architecture representation. On the basis of DARTS, Hundt\cite{hundt2019} proposed sharp DARTS, which is a more general, balanced, and consistent design. The closest concept to NASES is the approach proposed by Luo\cite{luo2018} in which an encoder and a decoder was to map neural architectures in a continuous space on gradient-based optimization and a predictor was used to achieve embedding accuracy.

\section{Methods}
    In this section, to elucidate the NASES procedure, we followed the aforementioned three dimensions: search strategy, search space, and performance estimation strategy.
    
\subsection{Search strategy}
    The search strategy is a search method for fast and accurate exploration of the space of neural architectures and involves techniques such as reinforcement learning, evolutionary algorithm, and gradient-based method. These are popular strategies in NAS.

    In the NAS with reinforcement learning, which is generally designed with two components of the controller network and child network (Figure 1), the controller network is used to control the child network architecture and generates string and the child network construct neural architecture by using the output of the controller in each NAS iteration. The controller network calculates the policy gradient to update the network by validating the performance of child network. Thus, severe penalty is imposed when performance is low. The controller that is constructed using a recurrent neural layer.
    
\begin{figure}[htb]
    \includegraphics[scale=0.7]{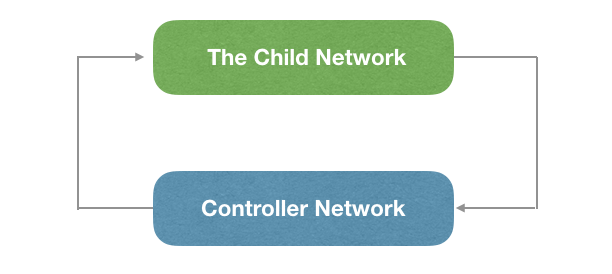}
    \centering 
    \caption{Overview of NAS with reinforcement learning. The child network receives information from the controller and generates a network architecture to evaluate date. The controller network receives a reward from the evaluation of the child network for updating network parameters.}
    \label{nas}
\end{figure}

\subsection{Search Space} 
    The main contributions of NASES is in the search space domain, which resolves the two aforementioned problems of noncontinuous and high-dimensional space in reinforcement learning; these problems lead to difficult optimization. NASES is similar to the general NAS procedure, which also includes the child network and controller network. The optimization of maximize accuracy is also used as a policy gradient method. However, our method differed from the general NAS procedure; first, we developed an architecture encoder as the controller network to control the architecture of the child network and projected origin architecture into architecture-embedding. Second, we devised the architecture decoder network, which decodes architecture-embedding from the architecture controller network to the origin architecture to ensure the child network can understand and generate network by using architecture-embedding (see Figure 2). That is, to alleviate these problems, the architecture decoder functions as a translator to translate low-dimension embedding into high-dimension vector for smooth child–controller network communication. 
    
    \begin{figure}[htb]
        \includegraphics[scale=0.7]{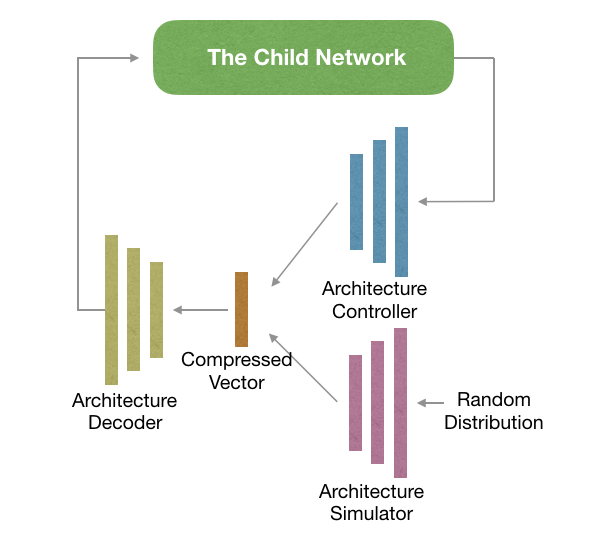}
        \centering 
        \caption{
            Overview of NAS on the embedding space. We used three functions namely architecture controller, architecture decoder, and architecture simulator, rather than the origin controller network for searching the neural architecture. Architecture controller received the origin architecture and generated architecture-embedding. Architecture decoder received the architecture-embedding and generated the origin architecture. Architecture simulator received random distribution and generated architecture-embedding.}
            \label{nases}
    \end{figure}
    
    \subsubsection{The Principal Functions of NASES.} 
        The NASES, has three principal functions. This section describes the functions of the architecture decoder, architecture simulator, and controller network.
    \paragraph{Architecture Decoder}

       To obtain architecture-embedding decoder, first we created an approximate of virtual distribution transformation, which projected the low-dimension space into high-dimension space. That is, we transformed architecture-embedding into origin architecture.
                
       $$f_{\theta}: \mathbb{R}^n \rightarrow  \mathbb{R}^m$$
                $$a= f_{\theta}(\hat{a})$$

        where $f_{\theta}$ is a decode function parameterized by $\theta$, $R^{n}$ is the architecture-embedding space, $R^{m}$ is the origin architecture space, $a$ is the set of  origin architecture, $\hat{a}$ is the set of  architecture-embedding. This function released the controller network architecture and could be projected on another space not bounded on the origin architecture space. This function is efficient and provides distribution transformers. To develop this approximator, an architecture simulator is required, which is discussed in the next subsection.        
        \paragraph{Architecture Simulator}
        An architecture simulator is an approach of the approximator of the virtual distribution transformation; its purpose creation of a function that simulates distribution to achieve architecture-embedding. Furthermore, the distribution of simulation is not limited to discrete or continuous space.

        $$s_{v}:U \rightarrow \mathbb{R}^{n}$$
        $$  \hat{a}= s_{v}(U) $$

        where $s_{v}$ is a function parametrized by $v$ and U is a uniform distribution.
        
        \paragraph{Obtaining the architecture encoder and architecture simulator}    
        To obtain the architecture encoder and simulator, we used the autoencoder network \cite{hinton2006}. The autoencoder network is a unsupervised algorithm for distribution transformation and dimension reduction. The autoencoder generates a representation by using the reduced encoding closest to its original input. Therefore, the architecture encoder and simulator were assembled in the autoencoder network. We pretrained an autoencoder network before training the controller, in which input space and target space had the same distribution. We used uniform distribution.
                
              
        \paragraph{Controller}
         A controller controls the child neural architecture. The neural architecture and hyperparameters of the controller are copied from the architecture simulator. The input is the child state of network, and the output is architecture-embedding.
                
          $$ g_{k}:\mathbb{R}^m \rightarrow \mathbb{R}^{n}$$
 
        where $g_{k}$ is an encode function parameterized by $k$, $R^{n}$ is the architecture-embedding space, $R^{m}$ is the origin architecture space, $A$ is the set of origin architecture, and $\hat{a}$  is the set of architecture-embedding. We did not retrain the controller on the NASES procedures again and only fine-tuned weights. The controller network can explore the architecture-embedding if the weights are initialized using the pretraining simulator of the autoencoder network because the pretraining simulator can project uniform distribution into architecture-embedding. Another advantage is fast searching because the controller network is not required to learn projecting uniform distribution into architecture-embedding again and focuses on searching architectures.
    \subsubsection{Child Model}
         The child model receives continuous vectors from the controller and generates neural architecture by using the pretrained decoder model. Here, we described the NASES mechanism for creating a network architecture. We use the 5 operators as ENAS\cite{Pham2018}: identity, 3 $\times$ 3 separable convolution, 5 $\times$ 5 separable convolution, 3 $\times$ 3 average pooling, 3$\times$3 maxpooling. The learning scheduler follows cosine annealing with $l_{max}$ = 0.05, $l_{min}$ = 0.001 and $T_0$ = 10 \cite{loshchilov2017}. Besides, each architecture search was only  run for 70 epochs on the search phrase, and final architecture is run for 630 epochs. The detailed is presented in Alg 1. More details are in Appendix A.


\begin{algorithm}[h] 
\caption{Neural Architecture Search in Embedding Space }
\label{alg::NASES}
\begin{algorithmic}[1]
\REQUIRE
uniform distribution $U$;
architecture simulator $s_v$;
architecture decoder $f_{\theta}$;
controller $g$;
child network $c$ 
number of optimization iterations $L$;
number of epochs in training phase $e_1$;
number of epochs in final phase $e_2$;
\STATE  $s_v$ and $f_{\theta}$ were assembled in an autoencoder network
\STATE Train the autoencoder network and evaluate its performance by uniform distribution $U$.
\STATE The controller $g$ weights initialized with architecture simulator pretrained weights $v$.
\FOR{$l$=1$,...,L$}
 \STATE Generate child network c by architecture $a$.
 \STATE Train and evaluate data child network $c$ in epochs $e_1$ by training and validation set.
 \STATE Compute reward $r$ from validation performance.
 \STATE Train controller network $g_v$ using reward $r$. 
 \STATE Project architecture $a$ into architecture-embedding $\hat{a}$ by using the controller network $g_v$.
 \STATE Decode architecture-embedding $\hat{a}$ into architecture $a$ by using the architecture decoder $f_{\theta}$.
\ENDFOR
\STATE Train and evaluate on the child network with the highest reward in epoch $e_2$ by training and testing set.
\end{algorithmic}
\end{algorithm}

\section{Experiments}
    We describe two stages of the experiment in this section. The first stage involves training an architecture decoder and architecture simulator network. The second stage involves applying the result of first stage to discover novel neural architectures for image classification on CIFAR-10 \cite{krizhevsky2009learning} by using NASES.
       
\subsection{First Stage: Pretraining Architecture Decoder and Simulator Network}

\subsubsection{Dataset} 
    In the first stage of NASES experiment, our goal was to map origin architecture to architecture-embedding. We assembled the architecture simulator and decoder in an autoencoder network and trained this autoencoder network instead of training the simulator and decoder. To mimic the origin architecture space. In this case, we sampled uniform distribution.
    
\subsubsection{Training Details}
    The optimization of the autoencoder network was achieved by using an Adam \cite{kingma2015} optimizer with a learning rate of 0.00001. During the training of the long short-term memory (LSTM) \cite{hochreiter1997} autoencoder network. In the network architecture, the details regarding the experimental procedures are as follows: a LSTM layer were used for the simulator and decoder networks; The architecture decoder network exhibited the same neural architecture and hyperparameters setting as the architecture simulator network. The loss function used a least square error.

\subsection{Second Stage: Image Classification on CIFAR-10}

\subsubsection{Dataset}
    The second stage of the experiment is a multiclass classification for assigning a class to the image object. The CIFAR-10 \cite{krizhevsky2009learning} data set consists of 60000 color images of 32 $\times$ 32 RGB in 100 classes. Each class has 6000 images with 5000 training data and 1000 testing data. Additionally, to achieve standardization and normalization, we applied only three standard data augmentation techniques: (1) Subtracting the mean and then dividing the answer by the standard deviation, which ensures that all variables have mean zero and standard deviation 1. (2) Centrally padding on training set to 40 $\times$ 40 and randomly cropping images back to 32 $\times$ 32. (3) Randomly flipping images horizontally.

\subsubsection{Settings}
    The spilled validation ratio was 0.9; we then randomly split 45000 and 5000 images for training and validation, respectively, in the neural searching procedure. Finally, we used 50000 images for training and 10000 images for testing when the NASES search procedure was complete. Architecture search procedure was run for 70 epochs on the search phase, and the final architecture was run for 630 epochs.
    
    The child network is described in the paragraph following the method section. The hyperparameters setting of the child network considers ENAS  \cite{Pham2018} as a reference. It was trained with Nesterov momentum, the momentum of 0.9 \cite{nesterov1983}. The learning schedule followed the learning rate decay with a cosine annealing for each batch ($l_{max}$ = 0.05, $l_{min}$ = 0.001, $T_0$ = 10) \cite{loshchilov2017}, batch size of 128, weight decay of 1e-4. We initialize $w$ with $He$ initialization\cite{he2015} in the child network. We designed a 15-layer convolutional architecture on micro search. By following the controller network perspectives, we took the pretrained parameters of an already trained model from the first phase experiment of the pretrained architecture simulator. The controller hyperparameters setting and neural architecture followed the first stage of the experiment.
\begin{table}[t]
\renewcommand\arraystretch{1.3} 
\begin{center}    
\setlength{\tabcolsep}{2.5mm}{
\begin{tabular}{c|ccccccc}
  \hline\hline
    \bfseries \textbf{Method}&\bfseries \textbf{N}&\bfseries \textbf{B}&\bfseries \textbf{F}&\bfseries \textbf{Params}&\bfseries \textbf{GPUs} &\bfseries \textbf{Error} 
  \\ \hline\hline
     DenseNet-BC\cite{huang2017} & -  & -  &-   & 25.6m & -  & 3.46&  \\  
     \hline 
     AmoebaNet-B + Cutout\cite{real2018} & 5  & 6  &128  & 34.9m & 3150  & 2.13 &\\  
     Hierarchical evolution\cite{Hanxiao2018} & 2  & 5  & 64   & 15.7m & 300  & 3.75&  \\
     PNAS +cutout\cite{liu2017}      & 3  & 5  &48  & 3.2m & 225  & 3.41 &\\   
     DARTS + cutout \cite{liu2018}   & 5  & 5  &36  & 4.6m & 4  & 2.83 &\\   
     ENAS \cite{Pham2018}             & 5  & 5  &36  & 4.6m & 0.45 & 3.54&  \\   
     ENAS + cutout \cite{Pham2018}   & 5  & 5  &36  & 4.6m & 0.45 & 2.89 &  \\     
     NAONet-WS \cite{luo2018}      & 5  & 5  &36  & 2.5m & 0.3 & 3.53&   \\
     NAONet-WS +cutout \cite{luo2018} & 5  & 5  &36  & 2.5m & 0.3 & 2.87&   \\  

     \hline   
     NASES             & 5  & 5  &40  &  4.7m & 0.15 & 3.45&  \\ 
     NASES + cutout    & 5  & 5  &40  & 4.7m & 0.15 & 2.85 &  \\     
 \hline\hline
\end{tabular}}
\end{center}
\caption{Performance and GPU computing time on macro search space of the NAS approach for class classification on CIFAR-10: First block represents the results of the manual network. The second represents the results of other excellent NAS works and the last blocks represent the results of the NASES final network.}
\end{table}

\subsubsection{Result}
    We ran the NASES procedure five times by using different random seeds on a single Nvidia V100 GPU, and NASES required approximately 3.5 hours to determine the final architecture for a NASES procedure. To determine the performance of the architecture, we evaluated the child network by using the final architecture network on the CIFAR-10 test dataset. Table 1 summarizes the performance of NASES and other NAS approaches by using the macro search algorithms.

    In Table 1, the approaches have been into three block. The first block of Table 1 presents the performances of manual network approaches, the second block represents the results of other excellent NAS works and the last blocks represent the results of the NASES final network; the NASES final architecture that achieves 3.45 error rate on the testing set uses only 4.7 million parameters in 3.5 hours, which is comparable with other NAS approaches. 
    
    For comparing more approaches and models, we added cutout randomly masks a square region in the images and the NASES final architecture can be improved to 2.85 error rate, which was better than approximately 4.6 million parameters used by the ENAS \cite{Pham2018} and NAS \cite{zoph2017}. NASES required approximately 0.15 GPU day to discover the final architecture.  The beneficial-performance and effectiveness of NASES was impressive even when only the architecture-embedding searching and pre-training controller were applied.
 
\section{Conclusion}
NAS with reinforcement learning is a powerful and novel framework for the automatic discovering process of neural architectures. Here, we designed a novel NAS framework, and this approach alleviated the two problems of noncontinuous and high-dimensional search space of NAS with reinforcement learning. We named this NAS framework NASES, in which the controller can be searched on embedding space by using the architecture decoder and architecture simulator. We achieved favorable results for image classification on CIFAR-10; the NASES exhibited efficient performance and high effectiveness.

\bibliography{iclr2020_conference}
\bibliographystyle{iclr2020_conference}
\newpage
\appendix
    
\section{The Child Model details}
     \subsection{The Order of the Blocks in Each Layer.}
            Performance was affected by the order of blocks in each layer. We applied the order of ReLu-conv-batchnorm \cite{ioffe2015}. Moreover, the kernel size of 1 $\times$ 1 convolution filters can be applied to change the dimensionality in the filter space. We applied the order of ReLu-conv-batchnorm to 1 $\times$ 1 kernel size convolution layers before the convolution layer, except for the first layer.     
    \subsection{Global Average Pooling.}
            We employed a trick into the NASES of the global average pooling \cite{lin2013} which is an operation  that calculates the average output of each feature map in the final convolution layer for reducing the number of parameters from the full connection layer.     

%
%
%
%

\end{document}